\documentclass[11pt]{article}

\usepackage[table]{xcolor}
\PassOptionsToPackage{hyperfootnotes=false}{hyperref}
\usepackage[preprint]{acl}

\usepackage{times}
\usepackage{latexsym}
\usepackage[T1]{fontenc}
\usepackage[utf8]{inputenc}
\usepackage{microtype}
\usepackage{inconsolata}
\usepackage{graphicx}

\usepackage{amssymb}
\usepackage{amsmath}
\usepackage{booktabs}
\usepackage{multirow}
\usepackage{tabularx}
\usepackage{arydshln}
\usepackage{enumitem}
\usepackage{url}

\newcommand{\hk}[1]{\textcolor{black}{#1}}

\title{IV-CoT: Implicit Visual Chain-of-Thought for Structure-Aware Text-to-Image Generation}

\author{
 \textbf{Zixuan Li\textsuperscript{1}\thanks{This work was done during the internship at Ant Group.}},
 \textbf{Haokun Lin\textsuperscript{1}\thanks{Corresponding author.}},
 \textbf{Yicheng Xiao\textsuperscript{3}},
 \textbf{Zhiwei Li\textsuperscript{1}},
 \textbf{Xinyang Song\textsuperscript{1}},
\\
 \textbf{Zelong Zheng\textsuperscript{1}},
 \textbf{Yong He\textsuperscript{2}},
 \textbf{Heng Yao\textsuperscript{2}},
 \textbf{Ke Ding\textsuperscript{2}},
 \textbf{Chao Yu\textsuperscript{2}},
\\
 \textbf{Chuan Yuan\textsuperscript{2}\footnotemark[2]},
 \textbf{Qi Li\textsuperscript{1}\footnotemark[2]},
 \textbf{Zhenan Sun\textsuperscript{1}}
\vspace{0.2cm}
\\
 \textsuperscript{1}NLPR, Institute of Automation, Chinese Academy of Sciences
\\
 \textsuperscript{2}Ant Group \quad
 \textsuperscript{3}The University of Hong Kong
\vspace{0.2cm}
\\ 
 {\small zixuan.li@nlpr.ia.ac.cn}
}

\begin{document}
\maketitle

\begin{abstract}
Unified multi-modal large language models (MLLMs) have achieved strong text-to-image generation quality, but still struggle with structure-aware prompt following, where object counts, spatial relations, attribute bindings, and coarse layouts must be preserved. We attribute this limitation in part to the entanglement of structural planning and appearance rendering within a single conditioning stream. To address this issue, we propose \textbf{Implicit Visual Chain-of-Thought (IV-CoT)}, a latent visual reasoning framework for query-conditioned image generation. IV-CoT decomposes the visual conditioning queries into a structural-to-semantic cascade, where structural queries first form a latent visual plan and semantic queries then render appearance conditioned on this plan. To guide the structural queries, we introduce training-only sketch supervision, which encourages them to capture structure from sketches without requiring sketch extraction or intermediate decoding at inference time. IV-CoT performs implicit CoT reasoning in a single forward pass and achieves superior results on GenEval and T2I-CompBench. Visualizations and analyses demonstrate that the learned structural and semantic queries play complementary roles in structure-aware generation.
\end{abstract}

\section{Introduction}

\begin{figure*}[htbp]
    \centering
    \includegraphics[width=\textwidth]{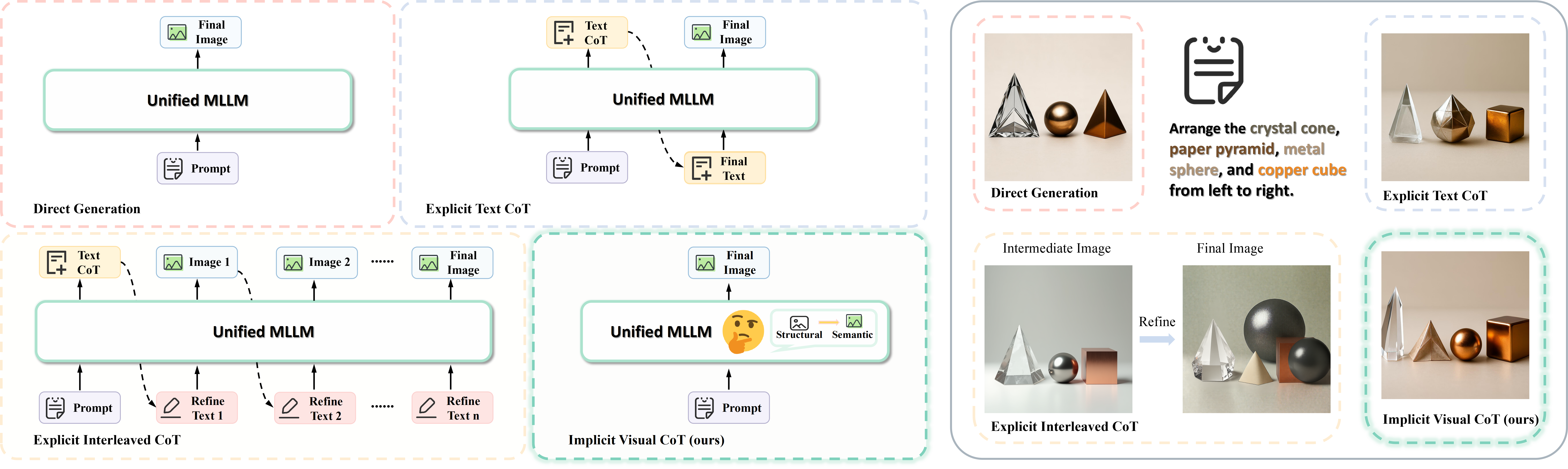}
    % \begin{subfigure}{0.48\textwidth}
    %     \centering
    %     \includegraphics[width=\textwidth]{figure/fig1_a.png}
    %     \caption{}
    %     \label{fig:taxonomy1}
    % \end{subfigure}
    % \hfill
    % \begin{subfigure}{0.48\textwidth}
    %     \centering
    %     \includegraphics[width=\textwidth]{figure/fig1_b.png}
    %     \caption{}
    %     \label{fig:taxonomy2}
    % \end{subfigure}

    % \caption{Comparison of reasoning paradigms for text-to-image generation. Unlike explicit textual or interleaved CoT methods, IV-CoT keeps the intermediate visual reasoning state in latent queries without decoding intermediate images.
    % \hk{not mentioned structural information here.}
    % }
    \caption{
Comparison of reasoning paradigms for text-to-image generation.
\textit{(Left)}: direct generation, explicit textual CoT, explicit interleaved CoT, and IV-CoT differ in where intermediate reasoning states are represented.
\textit{(Right)}: a structure-aware prompt-following example showing that IV-CoT better preserves the requested object layout without explicitly decoding intermediate images.
}
    \label{fig:taxonomy}
\end{figure*}

\hk{
Recent unified multi-modal large language models (MLLMs) have shown strong capabilities in generating realistic images from open-ended instructions~\citep{zhou2025transfusion,wu2025janus1,ma2025janusflow,cui2025emu3,xiao2025omnigen,xie2026showo2}. 
However, they still struggle with prompts that impose complex structural requirements~\citep{huang2023t2i,ghosh2023geneval,zhang2025itercomp,jiang2025draco}. 
When a prompt specifies multiple objects with distinct shapes, materials, attributes, and spatial arrangements, the model may produce visually plausible images while swapping attributes, omitting objects, or violating the requested layout. 
We refer to this setting as \textit{structure-aware prompt following}; an example is illustrated in Figure~\ref{fig:taxonomy}.
}

\hk{
Most unified MLLM-based generators convert the prompt into a visual conditioning stream through an understanding MLLM, where scene structure, object identity, attributes, and appearance details are compressed together~\citep{pan2025metaquery,wu2025openuni}. 
Such entangled conditioning makes it difficult for the generator to distinguish what should determine the scene structure from what should control visual appearance. 
Recent works therefore explore Chain-of-Thought (CoT) reasoning for image generation, using intermediate reasoning steps to better handle complex scenes~\citep{guo2025can,wang2025mcot_survey}.
}

\hk{
Existing CoT-based generation methods mainly follow two explicit paradigms.
\textbf{Explicit Textual CoT} generates intermediate verbal reasoning, scene descriptions, or numerical layouts before image synthesis~\citep{deng2025emerging,jiang2026t2i,tian2026unigen}. 
However, language alone has limited spatial bandwidth for representing continuous 2D geometry, object boundaries, relative scale, and occlusion.
\textbf{Explicit Interleaved CoT} incorporates visual intermediate states, such as masks, layouts, or draft images, to provide stronger structural cues~\citep{guo2025thinking,jiang2025draco,qin2025uni}. 
Yet this comes at the cost of explicit intermediate decoding, multi-stage pipelines, and potential error accumulation. 
These complementary limitations suggest that structure-aware generation needs an intermediate planning mechanism that preserves visual-spatial information while avoiding decoded reasoning states at inference time (Figure~\ref{fig:taxonomy}).}
% Rather than externalizing visual CoT as text, layouts, or decoded draft images, we ask whether structural planning can be internalized into a latent query cascade.

\hk{
Latent reasoning provides a natural alternative, where intermediate deliberation is carried by hidden states or soft thought tokens rather than explicit outputs~\citep{hao2024training,xu2025softcot,xu2025softcot++,ramji2026thinking}. 
Although related ideas have begun to appear in multimodal reasoning~\citep{pham2025multimodal,chen2025reasoning}, latent reasoning for text-to-image generation remains underexplored. 
The key question is: \textit{how should such latent states be organized before image rendering?}
% The key question is: \textit{what should such latent states represent, and how should they be organized before image rendering?}
We argue for a structure-first organization: the latent state should first anchor object boundaries, layout, and coarse shape, and then guide semantic appearance rendering. 
Without this structural scaffold, the generator may bind attributes to wrong objects, misplace spatial relations, or produce plausible details on incorrect structures.
}

We propose \textbf{Implicit Visual Chain-of-Thought (IV-CoT)}, a structure-first latent reasoning framework in the query space of a unified MLLM-DiT generator. 
IV-CoT internalizes structural planning into a \textbf{causal structural-to-semantic query cascade}: structural queries are placed before semantic queries and first encode a latent visual plan, including object form, count, layout, and coarse spatial relations; semantic queries then attend to this plan to render appearance and fine-grained details. 
In this way, the ``chain'' is realized as an ordered latent dependency from structural planning to semantic rendering, rather than an external sequence of text, layouts, or decoded images.

We further use sketches as training-only structural guidance through a two-stage training scheme.
Unlike sketch- or layout-conditioned generation methods that require external spatial controls at inference time, IV-CoT uses sketches only to shape latent structural queries during training.
In the first stage, sketch supervision encourages structural queries to encode contours, shapes, counts, and layouts while suppressing appearance factors such as color, texture, and lighting. 
In the second stage, IV-CoT is optimized for image generation with the structural objective retained as a regularizer, keeping structural queries aligned with the latent visual plan while semantic queries learn to render appearance details. 
At inference time, IV-CoT takes only the text prompt and produces structural and semantic queries in a single forward pass, without extracting sketches, decoding intermediate images, or generating explicit reasoning traces.
% At inference time, no sketch is extracted or decoded, and IV-CoT performs implicit CoT reasoning in a single forward pass, without intermediate image decoding or test-time reasoning traces. 
% Ablations confirm that neither more queries nor sketch supervision alone explains the improvement; the key is the combination of sketch-shaped structural queries and ordered structural-to-semantic dependency.
\hk{Our contributions are summarized as follows:}
% \vspace{-3mm}
\begin{itemize}[leftmargin=3mm, itemsep=0mm, topsep=-0.1mm]
\item 
We formulate Implicit Visual Chain-of-Thought, where intermediate visual planning is internalized in latent query representations rather than externalized as explicit textual or visual intermediate states.

\item 
We instantiate this formulation with a causal structural-to-semantic query cascade and training-only sketch supervision, which shape structural queries into latent visual plans while preserving text-only, single-pass inference.

\item Using the same OpenUni-L-1024 backbone, IV-CoT improves GenEval from 0.86 to 0.88 and T2I-CompBench from 0.5448 to 0.5743.
% with a pronounced gain on T2I spatial relation (0.4023 to 0.4734). 
Meanwhile, IV-CoT 
keeps single-pass inference and
achieves 9-15$\times$ lower latency than explicit CoT methods. 
Visualizations and cross-prompt recombination show that structural queries encode recoverable and manipulable latent visual plans. The relevant code will be released upon acceptance of the paper.
\end{itemize}

\section{Method}

\begin{figure*}[htbp]
    \centering
    \includegraphics[width=0.95\textwidth]{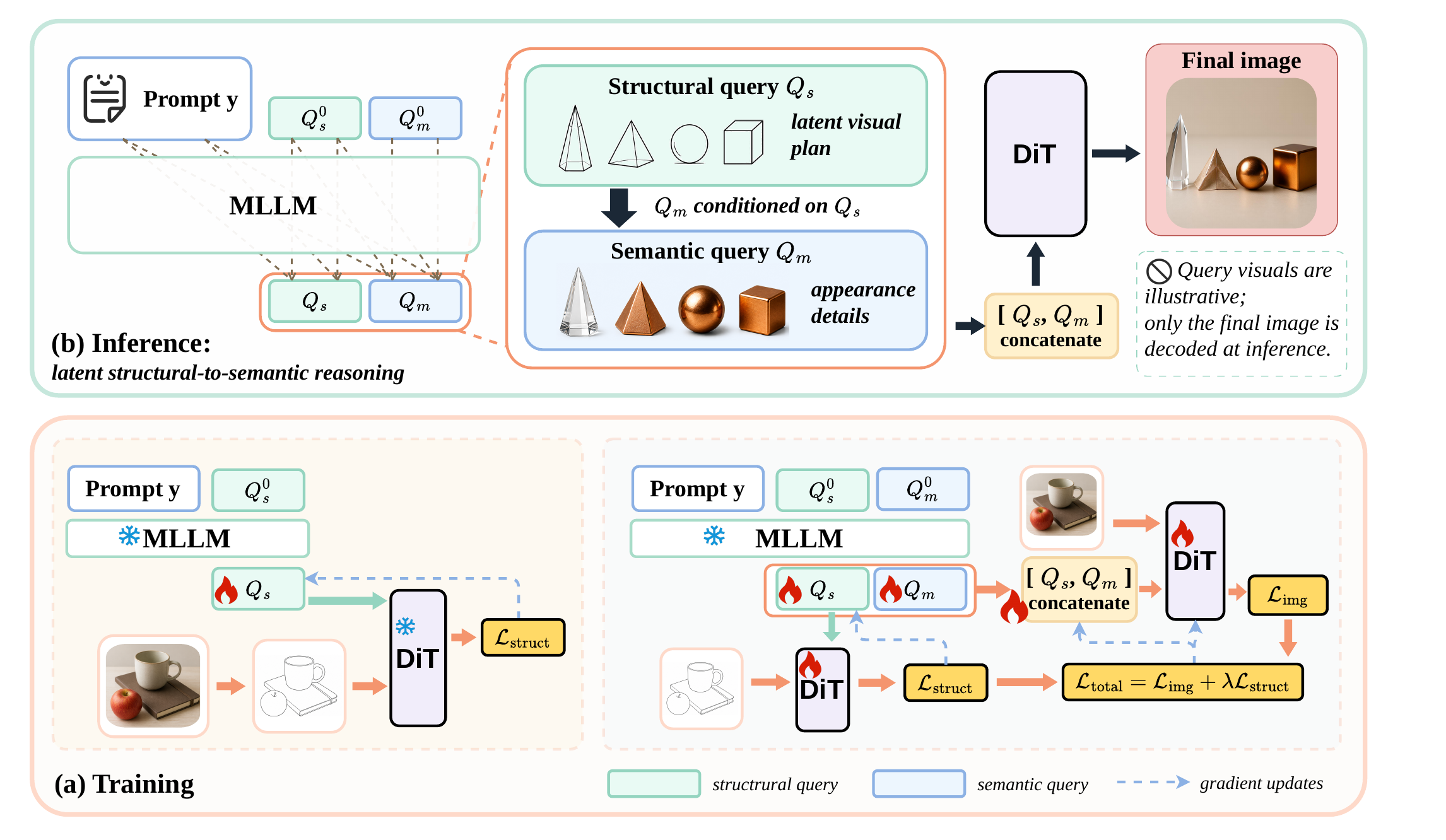}
    \caption{
Illustration of IV-CoT for text-to-image generation.
\textbf{(a)} During training, IV-CoT learns latent structural and semantic queries with structural and image-generation supervision.
\textbf{(b)} During inference, the MLLM performs latent structural-to-semantic reasoning, and the resulting queries guide the DiT to generate the final image.
No intermediate images are explicitly decoded; query visuals are shown only for illustration.
}
    \label{fig:main}
\end{figure*}

\hk{
We propose \textbf{Implicit Visual Chain-of-Thought (IV-CoT)}, a structure-first latent reasoning framework built upon a query-conditioned MLLM-DiT generator. 
Motivated by the difficulty of generation models in structure-aware prompt following (Figure~\ref{fig:taxonomy}), IV-CoT introduces an ordered structural-to-semantic query cascade that first encodes structural information and then performs semantic refinement. 
To enable such latent reasoning, we train the structural query inputs $\mathbf{Q}_{s}^{0}$ with a sketch-supervised objective and optimize image generation with structural regularization. 
This section first reviews the backbone architecture, followed by the proposed components and inference procedure.
}

% We propose \textbf{Implicit Visual Chain-of-Thought (IV-CoT)}, a structure-first latent reasoning framework, and instantiate it in a query-conditioned MLLM-DiT generator. The method consists of three components: an ordered structural-to-semantic query cascade, a sketch-supervised structural objective for shaping $\mathbf{Q}_{s}$, and image-generation training with structural regularization. This section first reviews the base architecture and then describes these components and the inference procedure.

\subsection{Query-Conditioned MLLM-DiT Generation}
\label{sec:method_prelim}

We build IV-CoT on a unified MLLM-DiT generation architecture~\citep{pan2025metaquery,wu2025openuni}. 
Given a text prompt $\mathbf{y}$ and a set of learnable visual query inputs 
$\mathbf{Q}_{0} \in \mathbb{R}^{N \times d}$, the MLLM produces continuous queries:
\begin{equation}
\mathbf{Q} = \Phi_{\mathrm{MLLM}}(\mathbf{y}, \mathbf{Q}_{0}),
\end{equation}
where $\mathbf{Q} \in \mathbb{R}^{N \times d}$ denotes the query states used for generation conditioning. 
In practice, a lightweight connector maps these query states into the conditioning space of the DiT; for simplicity, we use $\mathbf{Q}$ to denote the resulting conditioning sequence. 
The DiT conditions on $\mathbf{Q}$ to iteratively denoise a noisy image latent $\mathbf{z}_{t}$ and recover a clean image latent $\mathbf{z}_{x}$, which is decoded into the output image.

% In the standard formulation, all visual information is compressed into a single flat query sequence. The same queries must carry layout, object form, identity, attributes, color, and texture at once. This creates an entangled conditioning interface: the generator receives no explicit indication of which part of the condition should define the structure of the scene and which part should render appearance. IV-CoT modifies this interface by assigning different roles to different query groups and by enforcing an ordered dependency between them.
% \hk{
% In the MetaQuery formulation, visual information is compressed into a single flat query sequence. 
% As a result, the same sequence must jointly encode structural information, such as object layout, shape, identity, and attributes, as well as appearance information, such as color and texture. 
% This creates an entangled conditioning interface, where the DiT generator receives no explicit guidance on which conditioning factors should determine scene structure and which should control visual appearance. 
% Such entanglement can lead to structure-aware prompt-following errors, as illustrated in Figure~\ref{fig:compare}.
% This observation motivates IV-CoT, which restructures the conditioning interface by assigning distinct roles to different query groups and enforcing an ordered dependency from structural planning to semantic rendering.
% }
\hk{
In this formulation, visual information is compressed into a single flat query sequence, where structure-related factors, such as layout, shape and attributes, are entangled with appearance-related factors, such as color and texture. 
This provides the DiT generator with no explicit separation between structural planning and appearance rendering, leading to structure-aware prompt-following errors, as shown in Figure~\ref{fig:compare}.
IV-CoT addresses this limitation by assigning distinct roles to query groups and enforcing an ordered dependency from structure to semantics.
}

\subsection{Structural-to-Semantic Query Cascade}
\label{sec:method_cascade}

% IV-CoT partitions the learnable visual query inputs into two ordered groups: structural query inputs $\mathbf{Q}_{s}^{0} \in \mathbb{R}^{N_s \times d}$ and semantic query inputs $\mathbf{Q}_{m}^{0} \in \mathbb{R}^{N_m \times d}$. Given the text prompt $\mathbf{y}$, we feed the MLLM with the ordered sequence
% \begin{equation}
% [\mathbf{y}, \mathbf{Q}_{s}^{0}, \mathbf{Q}_{m}^{0}].
% \end{equation}
% The output hidden states at the structural and semantic query positions are used as the structural and semantic conditioning queries:
% \begin{equation}
% [\mathbf{Q}_{s}, \mathbf{Q}_{m}]
% =
% \Phi_{\mathrm{MLLM}}([\mathbf{y}, \mathbf{Q}_{s}^{0}, \mathbf{Q}_{m}^{0}]).
% \end{equation}

% Because the MLLM uses causal self-attention, this ordering induces a directional dependency between the two query groups. The structural queries are computed from the prompt and the structural query inputs, without access to the semantic query inputs. In contrast, the semantic queries are placed after the structural queries and can therefore attend to both the prompt and the structural query hidden states. This yields the structural-to-semantic dependency
% \begin{equation}
% \mathbf{Q}_{s} = \Phi_{s}(\mathbf{y}, \mathbf{Q}_{s}^{0}), \qquad
% \mathbf{Q}_{m} = \Phi_{m}(\mathbf{y}, \mathbf{Q}_{s}, \mathbf{Q}_{m}^{0}).
% \end{equation}

% The final conditioning sequence passed to the diffusion generator is
% \begin{equation}
% \mathbf{Q}_{\mathrm{IV-CoT}} = [\mathbf{Q}_{s}, \mathbf{Q}_{m}].
% \end{equation}

\hk{
IV-CoT partitions the learnable visual query inputs into two ordered groups: structural query inputs $\mathbf{Q}_{s}^{0} \in \mathbb{R}^{N_s \times d}$ and semantic query inputs $\mathbf{Q}_{m}^{0} \in \mathbb{R}^{N_m \times d}$. Given a text prompt $\mathbf{y}$, we feed the MLLM with the ordered sequence
\begin{equation}
[\mathbf{y}, \mathbf{Q}_{s}^{0}, \mathbf{Q}_{m}^{0}].
\end{equation}
The MLLM outputs are then divided into structural and semantic conditioning queries:
\begin{equation}
\mathbf{Q}_{s} = \Phi_{s}(\mathbf{y}, \mathbf{Q}_{s}^{0}), \qquad
\mathbf{Q}_{m} = \Phi_{m}(\mathbf{y}, \mathbf{Q}_{s}, \mathbf{Q}_{m}^{0}).
\end{equation}
Since the MLLM adopts causal self-attention, this ordering induces a one-way dependency between the two query groups. 
The structural queries are computed from the prompt and structural query inputs, without access to the semantic query inputs. 
In contrast, the semantic queries are placed after the structural queries and can therefore attend to both the prompt and the structural query states. 
This design establishes a structural-to-semantic cascade, encouraging the model to first form a latent visual plan and then render semantic appearance conditioned on it.
The final conditioning sequence passed to the diffusion generator is
}
\begin{equation}
\mathbf{Q}_{\mathrm{IV\mbox{-}CoT}} = [\mathbf{Q}_{s}, \mathbf{Q}_{m}].
\end{equation}

\subsection{Sketch-Supervised Structural Constraint}
\label{sec:method_struct}

\hk{
As shown in Figure~\ref{fig:main}(a), IV-CoT first uses a sketch-supervised structural constraint to guide the structural queries toward visual planning. 
Given a target image $\mathbf{x}$, we extract its sketch $\mathbf{s}$ with a fixed PiDiNet edge detector~\citep{su2021pixel} and encode it using the pretrained VAE encoder $\mathcal{E}$:
\begin{equation}
\mathbf{z}_{s} = \mathcal{E}(\mathbf{s}).
\end{equation}
Since sketches suppress appearance factors such as color, texture, and lighting while preserving contours, object shapes, counts, and coarse layouts, the sketch latent $\mathbf{z}_{s}$ defines a structure-focused clean latent for diffusion training.
}

\paragraph{Frozen-generator structural training.}

\hk{
We feed the ordered sequence $[\mathbf{y}, \mathbf{Q}_{s}^{0}]$ into the MLLM to obtain structural queries $\mathbf{Q}_{s}$, and use them to condition the DiT for sketch-latent denoising. 
Since $\mathbf{z}_{s}$ is encoded by the diffusion VAE, it lies in the same latent space as image latents and can be used as the clean latent for diffusion training.
During this stage, both the MLLM and DiT are frozen, and only the structural query inputs $\mathbf{Q}_{s}^{0}$ are optimized:
\begin{equation}
\mathcal{L}_{\mathrm{struct}}
=
\mathbb{E}_{\mathbf{z}_{s}, t, \epsilon}
\left[
\left\|
\epsilon - \epsilon_{\theta}(\mathbf{z}_{s,t}, t, \mathbf{Q}_{s})
\right\|_2^2
\right],
\end{equation}
where $\mathbf{z}_{s,t}$ denotes the noised sketch latent at diffusion step $t$. This frozen-generator design forces the structure required for sketch denoising to be encoded in $\mathbf{Q}_{s}$, rather than absorbed by adapting the MLLM or DiT.
}

\subsection{Semantic Rendering with Structural Regularization}
\label{sec:method_render}

\hk{
After structural training, we optimize IV-CoT for image generation. 
Given the ordered query sequence $[\mathbf{y}, \mathbf{Q}_{s}^{0}, \mathbf{Q}_{m}^{0}]$, the MLLM produces structural and semantic queries $[\mathbf{Q}_{s}, \mathbf{Q}_{m}]$, which are concatenated as the conditioning sequence for the diffusion generator. 
The image-generation objective is
\begin{equation}
\mathcal{L}_{\mathrm{img}}
=
\mathbb{E}_{\mathbf{z}_{x}, t, \epsilon}
\left[
\left\|
\epsilon - \epsilon_{\theta}(\mathbf{z}_{x,t}, t, [\mathbf{Q}_{s}, \mathbf{Q}_{m}])
\right\|_2^2
\right],
\end{equation}
where $\mathbf{z}_{x}=\mathcal{E}(\mathbf{x})$ denotes the clean image latent and $\mathbf{z}_{x,t}$ is its noised version at diffusion step $t$.
}

\hk{
To keep $\mathbf{Q}_{s}$ aligned with the sketch-induced visual plan, we retain the structural loss $\mathcal{L}_{\mathrm{struct}}$ as a regularizer during image-generation training. Unlike the first stage, where the generator is frozen to shape $\mathbf{Q}_{s}^{0}$, this stage optimizes the diffusion generator together with the query inputs. Gradients from $\mathcal{L}_{\mathrm{img}}$ update both structural and semantic query inputs, whereas $\mathcal{L}_{\mathrm{struct}}$ mainly regularizes the structural branch. This prevents $\mathbf{Q}_{s}$ from drifting toward appearance-only cues, while allowing $\mathbf{Q}_{m}$ to complete identity, color, material, and texture details conditioned on the structural queries.
The final objective is
\begin{equation}
\mathcal{L}_{\mathrm{total}}
=
\mathcal{L}_{\mathrm{img}}
+
\lambda \mathcal{L}_{\mathrm{struct}},
\end{equation}
where $\lambda$ controls the regularization strength.
}

\subsection{Inference}
\label{sec:method_infer}

% At inference time, 
\hk{As presented in Figure~\ref{fig:main} (b), during inference,}
IV-CoT uses the same full ordered query inputs as in semantic rendering, $[\mathbf{y}, \mathbf{Q}_{s}^{0}, \mathbf{Q}_{m}^{0}]$, and produces structural and semantic queries in one MLLM pass:
\begin{equation}
[\mathbf{Q}_{s}, \mathbf{Q}_{m}]
=
\Phi_{\mathrm{MLLM}}([\mathbf{y}, \mathbf{Q}_{s}^{0}, \mathbf{Q}_{m}^{0}]).
\end{equation}
No sketch is extracted or decoded. The DiT then performs standard diffusion decoding conditioned on the combined queries:
\begin{equation}
\hat{\mathbf{x}} = \Phi_{\mathrm{DiT}}(\mathbf{z}_{T}, [\mathbf{Q}_{s}, \mathbf{Q}_{m}]).
\end{equation}
Thus, IV-CoT introduces no explicit intermediate image generation, no reward-guided test-time search, and no additional visual decoding stage. The structural plan remains internal to the query sequence, while the visible output is produced by the usual image generation process.
\hk{
Notably, this differentiates IV-CoT from explicit textual or interleaved CoT methods, such as T2I-R1, GoT-R1, and TWIG, which typically require multiple reasoning steps. In contrast, IV-CoT achieves higher efficiency, as evidenced by the results in Table~\ref{table:latency}.
}
% \hk{add references and discuss our pros compared with previous work}

\section{Experiment}

\begin{table*}[t]

\centering
\small
\renewcommand{\arraystretch}{1.3}
\setlength{\tabcolsep}{1.22pt}

\newcolumntype{C}{>{\centering\arraybackslash}X}

\begin{tabularx}{\textwidth}{l *{7}{C} *{7}{C}} 
\toprule
\multirow{2}{*}{\textbf{Method}} & \multicolumn{7}{c}{\textbf{GenEval}} & \multicolumn{7}{c}{\textbf{T2I-CompBench}} \\
\cmidrule(lr){2-8} \cmidrule(lr){9-15}
& Single Obj. & Two Obj. & Count-ing & Colors & Pos-ition & Color Attri. & Over-all$\uparrow$ 
& Color & Shape & Texture & Spatial & Non-Spatial & Com-plex & Over-all$\uparrow$ \\
\midrule
\multicolumn{15}{l}{\color{gray}{\textit{Unified models}}} \\
\hdashline
Janus-Pro-7B & 0.98 & 0.85 & 0.56 & 0.89 & 0.77 & 0.64 & 0.78 & 0.6359 & 0.3528 & 0.7243 & 0.3378 & 0.3085 & 0.3559 & 0.4525 \\
Emu3 & 0.98 & 0.71 & 0.34 & 0.81 & 0.17 & 0.21 & 0.54 & 0.7544 & 0.5706 & 0.7164 & - & - & - & - \\
Show-o & 0.95 & 0.52 & 0.49 & 0.82 & 0.11 & 0.28 & 0.68 & 0.5600 & 0.4100 & 0.4600 & 0.2000 & 0.3000 & 0.2900 & 0.3700 \\
MetaQuery-XL & - & - & - & - & - & - & 0.80 & - & - & - & - & - & - & - \\
OpenUni-L-1024 & 0.99 & 0.92 & 0.76 & 0.91 & 0.82 & 0.77 & 0.86 & 0.8216 & 0.6117 & 0.7308 & 0.4023 & 0.3102 & 0.3923 & 0.5448 \\
BAGEL & 0.99 & 0.94 & 0.81 & 0.88 & 0.64 & 0.63 & 0.82 & 0.8027 & 0.5685 & 0.7021 & 0.3488 & 0.3101 & 0.3824 & 0.5191 \\
TUNA-2 & 0.99 & 0.96 & 0.80 & 0.91 & 0.84 & 0.76 & \underline{0.87} & - & - & - & - & - & - & - \\
\midrule
\multicolumn{15}{l}{\color{gray}{\textit{Unified models with explicit CoT}}} \\
\hdashline
T2I-R1$^{*}$ & 0.99 & 0.91 & 0.53 & 0.91 & 0.76 & 0.65 & 0.79 & 0.8130 & 0.5852 & 0.7243 & 0.3378 & 0.3090 & 0.3993 & 0.5281 \\
GoT-R1$^{*}$ & 0.99 & 0.94 & 0.50 & 0.90 & 0.46 & 0.68 & 0.75 & 0.8139 & 0.5549 & 0.7339 & 0.3306 & 0.3169 & 0.3944 & 0.5241 \\
TWIG-RL$^{*}$ & - & - & - & - & - & - & - & 0.8249 & 0.6128 & 0.7319 & 0.3406 & 0.3199 & 0.5445 & \underline{0.5624} \\
Uni-CoT & 0.99 & 0.96 & 0.84 & 0.92 & 0.57 & 0.71 & 0.83 & - & - & - & - & - & - & - \\
Draco & 1.00 & 0.99 & 0.81 & 0.91 & 0.70 & 0.76 & 0.86 & - & - & - & - & - & - & - \\
\midrule
\rowcolor{blue!6}
IV-CoT (Ours) & 1.00 & 0.96 & 0.78 & 0.92 & 0.86 & 0.79 & \textbf{0.88} & 0.8542 & 0.6296 & 0.7550 & 0.4734 & 0.3199 & 0.4136 & \textbf{0.5743} \\
\bottomrule
\end{tabularx}
\caption{Model performance comparison on GenEval and T2I-CompBench. Best results are in bold and second-best results are underlined. Methods marked with $^{*}$ use prompts from the T2I-CompBench training split during training.}
\label{table:main_result}
\end{table*}

In this section, we conduct comprehensive experiments to answer the following Research Questions (RQs):
% }

% leftmargin=5mm, 
\begin{enumerate}[itemsep=0mm, topsep=-0.5mm]
    \item[\textbf{RQ1:}] Does IV-CoT improve compositional and structure-aware prompt following while preserving inference efficiency?
    \item[\textbf{RQ2:}] Are structural supervision and query cascade both necessary?
    \item[\textbf{RQ3:}] Do structural queries encode recoverable visual plans and actively influence generated structure?
    \item[\textbf{RQ4:}] Does query separation enable zero-shot structure-appearance recombination?
\end{enumerate}

\subsection{Setup}
We instantiate IV-CoT on OpenUni-L to isolate the effect of the proposed query organization and structural supervision. We fine-tune IV-CoT on BLIP3-o~\citep{chen2025blip3}, ShareGPT-4o~\citep{chen2025sharegpt}, and Echo-4o~\citep{ye2025echo}.
During structural training, we extract sketches from training images using a fixed PiDiNet edge detector as training-only supervision. 
No sketch is required at inference time.
Implementation details are provided in Appendix~\ref{app:impl}.

\paragraph{Baselines.}
We compare IV-CoT with two groups of methods. The first group includes unified multimodal generation models, such as Janus-Pro~\citep{chen2025janus}, Emu3~\citep{wang2024emu3}, Show-o~\citep{chen2026show}, MetaQuery-XL~\citep{pan2025metaquery}, OpenUni-L-1024~\citep{wu2025openuni}, BAGEL~\citep{deng2025emerging}, and TUNA-2~\citep{liu2026tuna}. The second group includes unified generation models with explicit CoT or reasoning-enhanced generation, including T2I-R1~\citep{jiang2026t2i}, GoT-R1~\citep{duan2025got}, TWIG-RL~\citep{guo2025thinking}, Uni-CoT~\citep{qin2025uni}, and Draco~\citep{jiang2025draco}. 
We instantiate IV-CoT on OpenUni-L-1024 to isolate design effect.
% We instantiate IV-CoT on OpenUni-L-1024, which provides the same backbone reference for measuring the effect of the proposed query organization and structural supervision.

\paragraph{Evaluation Benchmarks.}
We conduct the main evaluation on two widely used text-to-image generation benchmarks,
% used structure-aware prompt following on 
GenEval~\citep{ghosh2023geneval} and T2I-CompBench~\citep{huang2023t2i}. 
% GenEval measures compositional generation across object existence, counting, color, position, and attribute binding. T2I-CompBench evaluates open-world compositional prompt following along dimensions such as color, shape, texture, spatial relation, non-spatial relation, and complex composition. For same-backbone variants, we use the same evaluation protocol and report the official benchmark scores.

\begin{figure*}[htbp]
    \includegraphics[width=\textwidth]{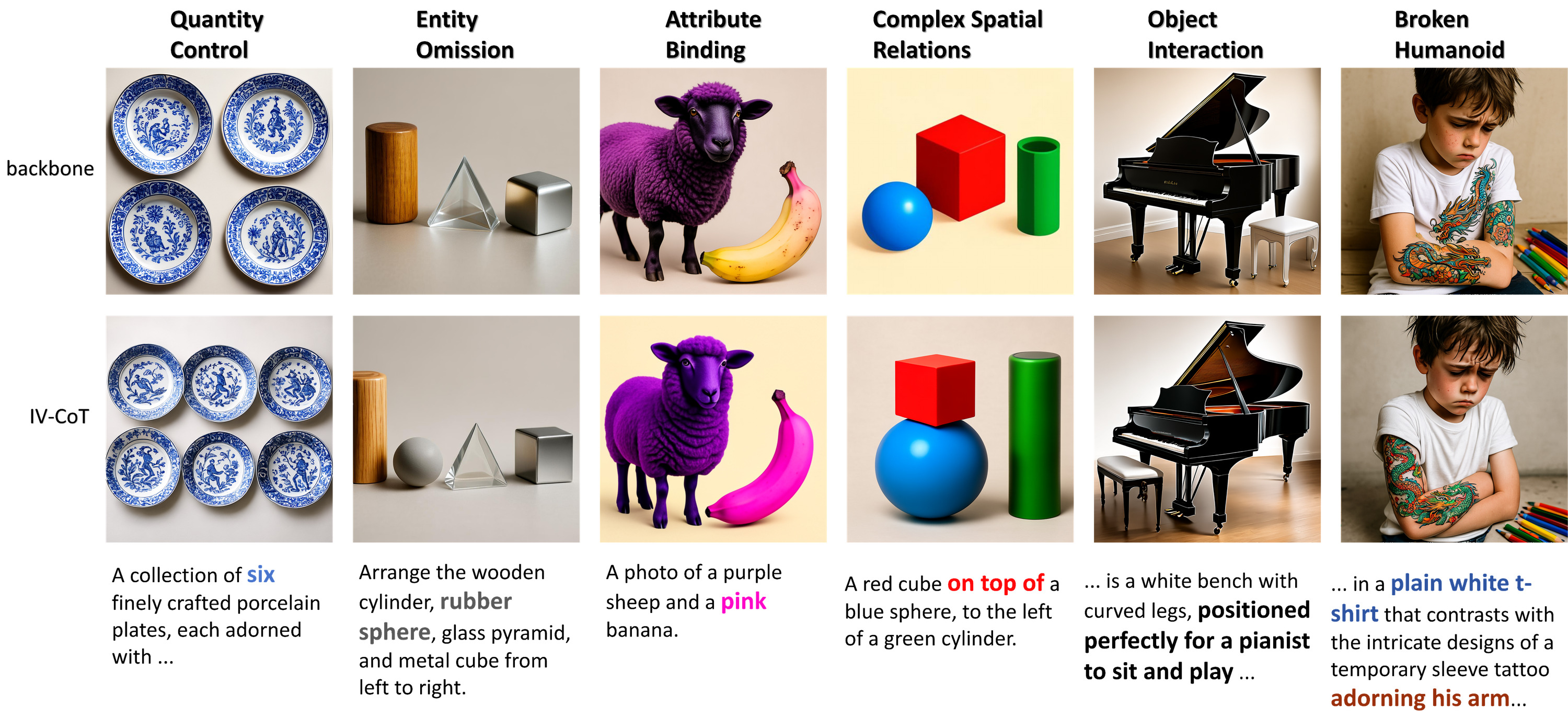}
    \caption{Qualitative comparison on structure-aware prompts. Each column corresponds to one prompt, with highlighted words indicating the target structural or attribute constraint. 
    % Compared with the baseline, 
    IV-CoT better maintains object count, spatial arrangement, attribute binding, and coarse object geometry while preserving visual quality.}
    \label{fig:compare}
    % \vspace{-5pt}
\end{figure*}

% To rigorously evaluate our framework, we design experiments to answer four Research Questions (RQs):
% \begin{enumerate}
%     \item[RQ1] Does Latent Visual CoT achieve state-of-the-art compositional performance?
%     \item[RQ2] Is there a strict causal dependency within the Latent CoT, and does the structural query genuinely serve as an indispensable topological anchor?
%     \item[RQ3] Can we open the "black box" of the implicit reasoning process and visually verify the decoupled roles of drafting and rendering?
%     \item[RQ4] To what extent does the framework achieve orthogonal disentanglement of form and texture in the latent space?
% \end{enumerate} 

\subsection{Main Results (RQ1)}

% \paragraph{Performance.}

% Table~\ref{table:main_result} reports the main results on GenEval and T2I-CompBench. 
% Compared with the same OpenUni-L-1024 backbone, IV-CoT improves the overall score from 0.86 to 0.88 on GenEval and from 0.5448 to 0.5743 on T2I-CompBench. 
% The gains are especially visible on structure-sensitive dimensions, such as position and color attribution in GenEval, and spatial relation, shape, texture, and color in T2I-CompBench. 
% This trend supports our central hypothesis that organizing conditioning queries into a structure-first latent plan helps the generator follow prompts that require accurate layout, object geometry, and attribute grounding.

% Compared with recent unified generation models and explicit CoT-based generation methods, IV-CoT achieves the best overall scores on both benchmarks among the reported methods. 
% Notably, this improvement does not rely on decoded intermediate images or test-time reasoning traces, suggesting that implicit visual CoT can provide the benefits of visual planning while keeping inference in a standard single-pass query-conditioned generation pipeline.

\paragraph{Performance.}
Table~\ref{table:main_result} reports the main results on GenEval and T2I-CompBench. 
Compared with OpenUni-L-1024, IV-CoT improves the overall score from 0.86 to 0.88 on GenEval and from 0.5448 to 0.5743 on T2I-CompBench. 
The gains are particularly evident on structure-sensitive dimensions, including position and color attribution in GenEval, as well as spatial relation, shape, texture, and color in T2I-CompBench. 
These results support the effectiveness of our structure-first, semantic-second pipeline for structure-aware prompt following.
Compared with recent unified generation models and explicit CoT-based generation methods, IV-CoT achieves the best overall scores on both benchmarks. 
Notably, IV-CoT performs inference in a single forward pass, without decoded intermediate image or test-time reasoning.

\paragraph{Generation Samples.}
\hk{
We provide a qualitative comparison with the OpenUni on structure-aware prompts in Figure~\ref{fig:compare}. 
The baseline often generates visually plausible images but fails to satisfy key structural constraints, such as object count, spatial placement, and attribute binding. 
In contrast, IV-CoT better preserves the specified visual organization while maintaining comparable image quality, further demonstrating the effectiveness of latent visual reasoning for structure-aware generation.
}
Additional generation samples are provided in Appendix~\ref{appendix:additional_samples}.

% Figure~\ref{fig:compare} further illustrates this effect qualitatively. 
% The baseline often produces visually plausible images but misses key structural constraints, such as object number, spatial placement, or attribute binding. 
% In contrast, IV-CoT better preserves the specified visual organization while maintaining comparable image quality, consistent with the quantitative improvements in Table~\ref{table:main_result}.

% \begin{figure}[htbp]
%     \includegraphics[width=0.43\textwidth]{figure/efficiency_bubble.png}
%     \caption{Efficiency-performance comparison on T2I-CompBench. The x-axis reports inference time, the y-axis reports T2I-CompBench overall score, and bubble size indicates the reported backbone/component scale. IV-CoT achieves the best score with substantially lower latency than explicit CoT-based generation methods.}
%     \label{fig:latency}
% \end{figure}
\paragraph{Inference efficiency.}
We further compare inference efficiency with explicit CoT-based generation methods in Table~\ref{table:latency}. IV-CoT achieves the highest T2I-CompBench overall score while requiring only 1.693 seconds per sample, which is 9.01$\times$, 9.89$\times$, and 14.98$\times$ lower latency than T2I-R1, GoT-R1, and TWIG-RL, respectively.
Details of latency measurement are provided in Appendix~\ref{app:latency}.
\begin{table}[t]
\centering
\scriptsize
\renewcommand{\arraystretch}{1.08}
\setlength{\tabcolsep}{1.5pt}
\begin{tabularx}{0.48\textwidth}{@{}*{5}{>{\centering\arraybackslash}X}@{}}
\toprule
\textbf{Method} & \textbf{Params} & \textbf{Time$\downarrow$} & \textbf{Rel.$\downarrow$} & \textbf{T2I-Comp$\uparrow$} \\
\midrule
T2I-R1 & 7B & 15.261 & 9.01$\times$ & 0.5281 \\
GoT-R1 & 7B & 16.744 & 9.89$\times$ & 0.5241 \\
TWIG-RL & 7B & 25.362 & 14.98$\times$ & 0.5624 \\
\rowcolor{blue!6}
IV-CoT (Ours) & 3.6B & \textbf{1.693} & \textbf{1.00$\times$} & \textbf{0.5743} \\
\bottomrule
\end{tabularx}
\caption{Efficiency comparison with explicit CoT generation methods. 
Rel. denotes latency relative to IV-CoT.}
\label{table:latency}
\vspace{-10pt}
\end{table}

\subsection{Ablation Study (RQ2)}
\label{sec:rq2_ablation}

\hk{
RQ2 studies whether the gains of IV-CoT arise from its structural-to-semantic design rather than simpler alternatives. All ablation variants use the same training data, training budget, and inference settings.
\textbf{Base} denotes OpenUni-L-1024, and \textbf{Base + More Queries} matches the number of visual queries used by IV-CoT to test the effect of query capacity.
\textbf{Flat Sketch Aux} applies the same sketch-based structural supervision to a flat query set, while \textbf{Parallel Two-Query} separates structural and semantic queries but generates them independently before concatenation. 
These variants isolate the effect of the ordered query cascade. 
\textbf{IV-CoT w/o Structural Constraint} keeps the cascade but removes sketch supervision on $\mathbf{Q}_{s}$, whereas \textbf{Full IV-CoT} uses both sketch-supervised structural training and the structural-to-semantic query cascade.
}

% RQ2 examines whether the improvement comes from the full structural-to-semantic design, rather than from simpler alternatives. \textbf{Base} denotes OpenUni-L-1024 under our evaluation setting. \textbf{Base + More Queries} increases the number of visual queries to match IV-CoT, testing whether the gain is merely due to larger query capacity. \textbf{Flat Sketch Aux} applies the same structural supervision to a flat query set without separating structural and semantic queries, testing whether sketch supervision alone is sufficient. \textbf{Parallel Two-Query} separates structural and semantic queries but generates them independently and only concatenates them before the diffusion generator, testing whether query partition alone explains the improvement. \textbf{IV-CoT w/o Structural Constraint} keeps the structural-to-semantic cascade but removes structural supervision on $\mathbf{Q}_{s}$. \textbf{Full IV-CoT} uses both structural supervision and the structural-to-semantic query cascade.

\begin{table*}[t]
\centering
\small
\renewcommand{\arraystretch}{1.15}
\setlength{\tabcolsep}{3pt}
\begin{tabular}{lcccccc}
\toprule
\textbf{Method} & \textbf{\#Queries} & \textbf{Query Split} & \textbf{Structural Supervision} & \textbf{Cascade} & \textbf{GenEval} & \textbf{T2I-CompBench} \\
\midrule
Base & 256 &  &  &  & 0.860 & 0.5448 \\
Base + More Queries & 512 &  &  &  & 0.869 & 0.5507 \\
Flat Sketch Aux & 512 &  & \checkmark &  & 0.866 & 0.5496 \\
Parallel Two-Query & 512 & \checkmark & \checkmark &  & 0.869 & 0.5562 \\
IV-CoT w/o Structural Constraint & 512 & \checkmark &  & \checkmark & 0.861 & 0.5573 \\
Full IV-CoT & 512 & \checkmark & \checkmark & \checkmark & 0.885 & 0.5743 \\
\bottomrule
\end{tabular}
\caption{Ablation study of IV-CoT. All ablation variants are trained with the same data mixture and training budget.}
\label{table:ablation}
\vspace{-5pt}
\end{table*}

% \hk{
Table~\ref{table:ablation} shows that Full IV-CoT performs best among all controlled variants.
Increasing the number of queries yields only limited gains, indicating that the improvement is not simply due to larger query capacity. 
Flat Sketch Aux and Parallel Two-Query remain below Full IV-CoT, suggesting that neither sketch supervision nor query partition alone fully explains the gain. 
Removing the structural constraint also weakens performance, confirming the need to explicitly guide $\mathbf{Q}_{s}$ toward structural planning.
These results indicate that structural supervision and the structural-to-semantic cascade are complementary: structural supervision shapes $\mathbf{Q}_{s}$ into a latent visual plan, while the cascade allows $\mathbf{Q}_{m}$ to render appearance conditioned on it.
% }

% Table~\ref{table:ablation} shows that Full IV-CoT achieves the best performance among all controlled variants. Increasing the number of queries improves over the base model, indicating that additional query capacity is helpful but insufficient. Flat Sketch Aux and Parallel Two-Query remain below Full IV-CoT, suggesting that sketch supervision or query partition alone does not fully account for the gain. Removing the structural constraint also weakens performance, showing that the cascade needs explicit structural supervision to make $\mathbf{Q}_{s}$ serve as a reliable latent visual plan. Overall, the results indicate that structural supervision and the structural-to-semantic cascade are complementary: the former shapes $\mathbf{Q}_{s}$ into a structural plan, while the latter lets $\mathbf{Q}_{m}$ render appearance conditioned on that plan.

\begin{figure}[!t]
\centering
\includegraphics[width=0.45\textwidth]{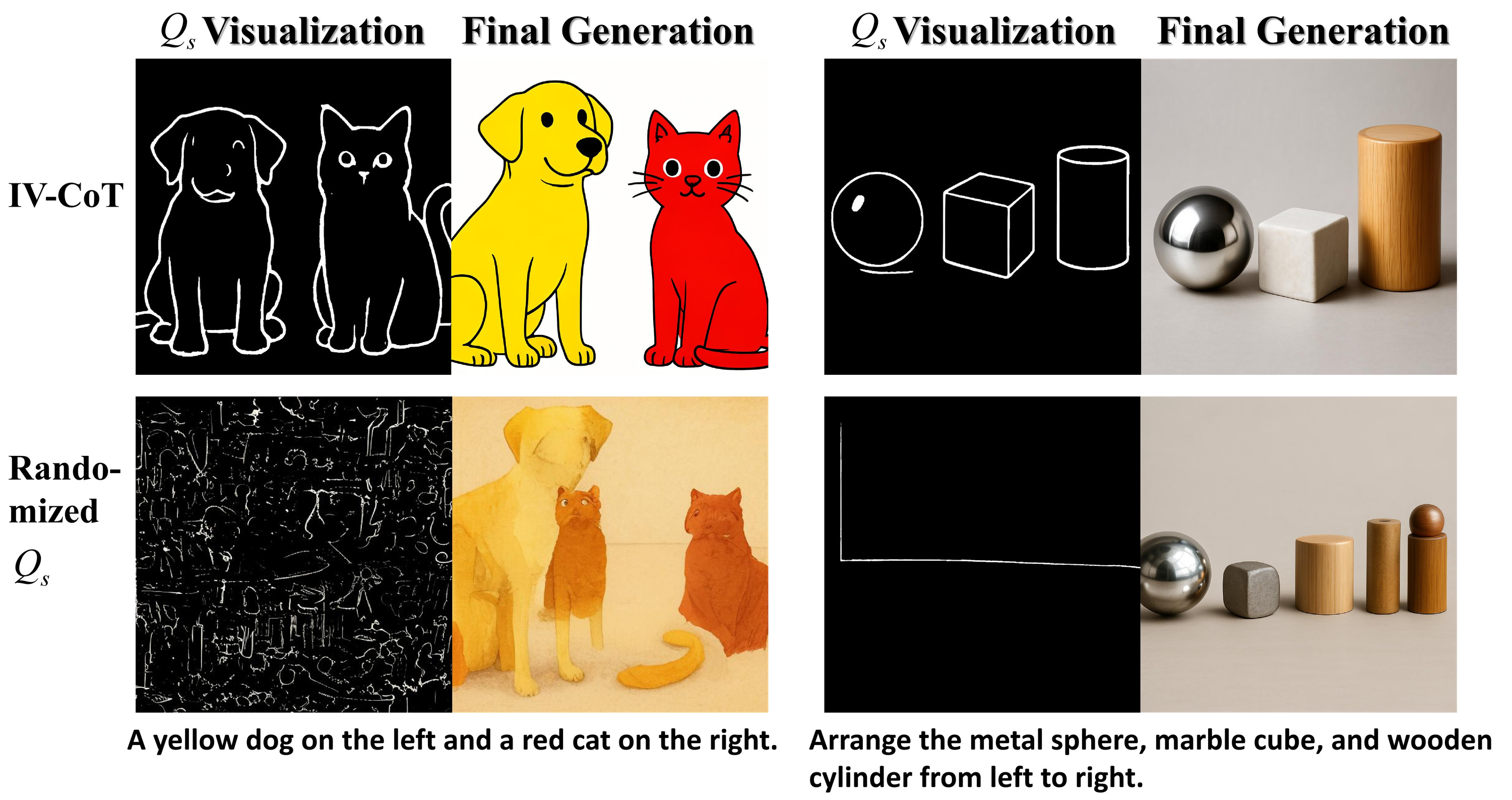}
    \caption{Visualization and perturbation of structural queries. 
    $\mathbf{Q}_{s}$ visualization reveals sketch-like latent visual plans that preserve object contours, counts, and coarse layouts. 
    Perturbing $\mathbf{Q}_{s}$ with random queries before diffusion decoding disrupts the generated layout and object geometry, indicating that $\mathbf{Q}_{s}$ acts as an active structural condition. 
    % rather than only an auxiliary training target. 
    % The decoded sketches are used only for analysis and are not produced during inference.
    }
    \label{fig:sketch}
    \vspace{-5pt}
\end{figure}

% \subsection{Opening the Black Box: Visualizing Latent Thoughts(RQ3)}
% \hk{What are latent thoughts? First Appear?}
\subsection{Opening the Black Box: Interpreting Latent Visual Plans (RQ3)}

RQ3 examines whether structural queries encode latent visual plans and whether the generator uses structural and semantic queries in different ways. 
% We provide two diagnostic analyses: query decoding and perturbation in Figure~\ref{fig:sketch}, and cross-attention proportion analysis in Figure~\ref{fig:heatmap}.

\paragraph{Query decoding and perturbation.}

% \hk{
We first examine whether structural queries encode recoverable visual plans in Figure~\ref{fig:sketch}. 
We use $\mathbf{Q}_{s}$ as the conditioning input to the DiT and visualize the corresponding sketch-domain generations. 
The results exhibit sketch-like structures that capture object shapes and scene layouts. 
In contrast, replacing $\mathbf{Q}_{s}$ with randomly initialized queries before diffusion decoding substantially disrupts object layout, contours, and coarse shape. 
This suggests that $\mathbf{Q}_{s}$ is actively used as a structural condition for image synthesis, rather than merely satisfying the sketch auxiliary loss.
These visualizations are used only for analysis; IV-CoT does not decode intermediate sketches during actual inference.
% }

% We first examine whether structural queries encode recoverable visual plans. As shown in Figure~\ref{fig:sketch}, decoding $\mathbf{Q}_{s}$ produces sketch-like structural representations that preserve object contours, coarse shapes, counts, and rough spatial layouts. These decoded sketches are used only for analysis; during inference, IV-CoT does not decode any intermediate sketch or visual state.

% We then perturb $\mathbf{Q}_{s}$ by replacing it with randomly initialized query vectors before diffusion decoding, while keeping the remaining conditioning unchanged. Perturbing $\mathbf{Q}_{s}$ substantially disrupts the generated structure, including object layout, contours, and coarse shape. This suggests that $\mathbf{Q}_{s}$ is not merely optimized to satisfy the sketch auxiliary loss, but is actively used by the generator as a structural condition for image synthesis.

\paragraph{Cross-attention proportion analysis.}

\begin{figure}[!t]
\centering
    \includegraphics[width=0.48\textwidth]{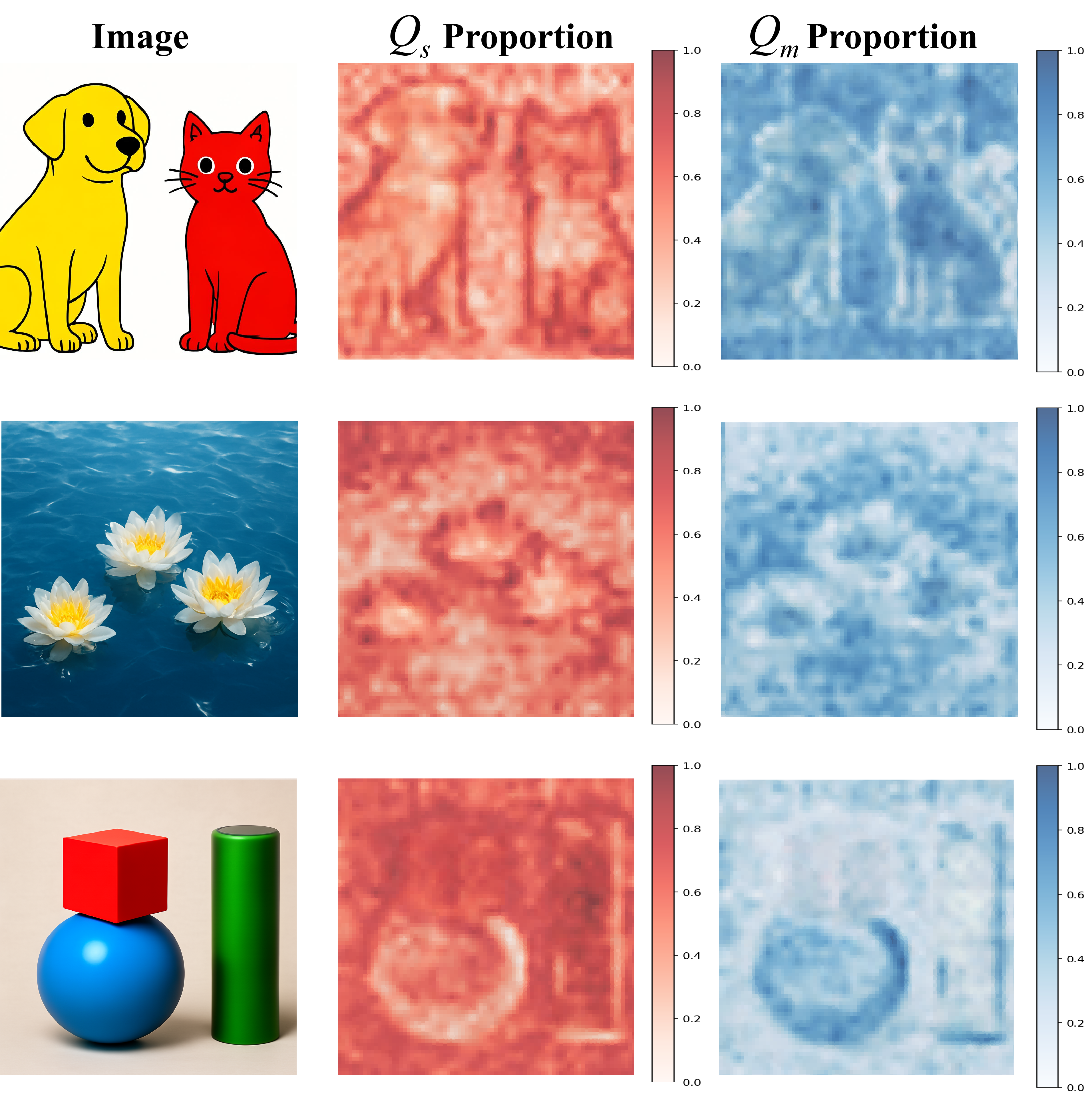}
    % \caption{Relative cross-attention proportion maps for structural and semantic queries. For each example, the image is shown alongside the normalized attention proportion assigned to $\mathbf{Q}_{s}$ and the complementary proportion assigned to $\mathbf{Q}_{m}$. Higher $\mathbf{Q}_{s}$ proportions tend to align with contours and spatial boundaries, while $\mathbf{Q}_{m}$ is more broadly distributed over appearance-related regions. These maps indicate soft specialization rather than a hard separation between the two query groups.}
    \caption{Relative cross-attention proportion maps. 
Each row shows the generated image, the normalized proportion assigned to structural queries $\mathbf{Q}_{s}$, and the complementary proportion assigned to semantic queries $\mathbf{Q}_{m}$. 
Structural queries receive higher relative attention around contours and spatial boundaries.}
    
    \label{fig:heatmap}
    \vspace{-10pt}
\end{figure}

\hk{
We further analyze how the diffusion generator allocates attention to structural and semantic queries during rendering. 
For each spatial latent position, we compute the relative cross-attention proportion assigned to $\mathbf{Q}_{s}$ and $\mathbf{Q}_{m}$, averaged over selected middle denoising steps. 
Since the two query groups have the same size, the comparison is not biased by group cardinality. 
Details and stage- and layer-wise visualizations are provided in Appendix~\ref{appendix:attention}.
}
% \vspace{-10mm}

\hk{
Figure~\ref{fig:heatmap} shows that $\mathbf{Q}_{s}$ receives higher relative attention around object contours, boundaries, and coarse spatial structures, while $\mathbf{Q}_{m}$ is more broadly activated over object interiors and appearance-related regions. This does not indicate a hard separation, but suggests a soft functional specialization: structural queries provide spatial guidance, whereas semantic queries support appearance rendering conditioned on the structural plan.
}

\begin{figure}[!t]
    \includegraphics[width=0.45\textwidth]{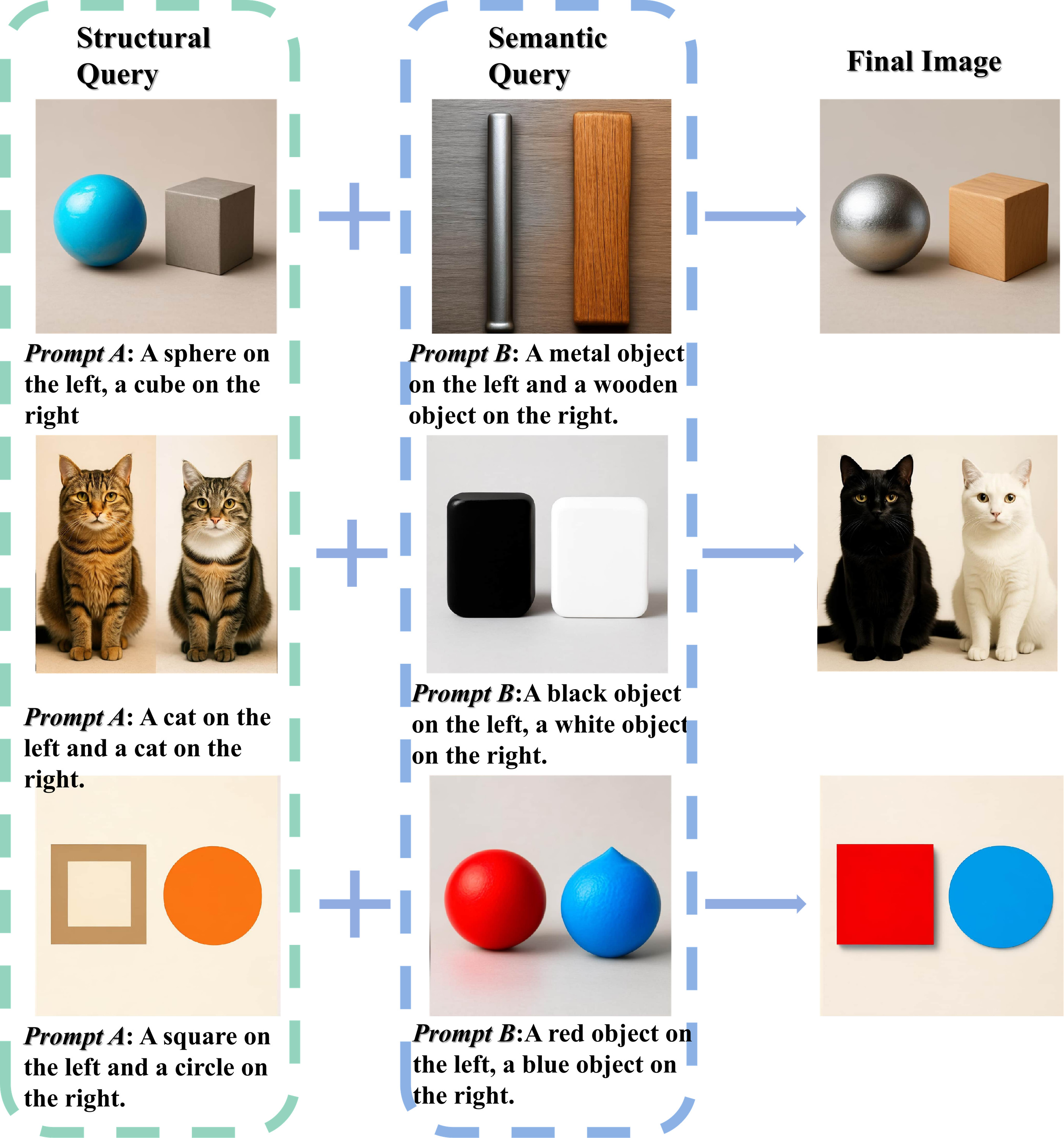}
    % \caption{Cross-prompt structure-appearance recombination. In each row, structural queries from Prompt A are combined with semantic queries from Prompt B and decoded by the diffusion generator. The mixed output preserves the coarse layout and object configuration from Prompt A while adopting appearance, material, or color attributes from Prompt B, illustrating controllable manipulation in the latent query space.}
    \caption{Cross-prompt structure-appearance recombination. 
Structural queries from Prompt A are combined with semantic queries from Prompt B. 
The mixed outputs tend to preserve the coarse layout from Prompt A while adopting appearance attributes from Prompt B, suggesting partial controllability in the latent space.}

    \label{fig:frankenstein}
    \vspace{-5pt}
\end{figure}

\subsection{Zero-Shot Structure-Appearance Recombination (RQ4)}

\hk{
We further examine whether the learned query separation provides controllable handles in the latent space. 
Given two prompts $y_A$ and $y_B$, IV-CoT produces query pairs $(\mathbf{Q}_{s}^{A}, \mathbf{Q}_{m}^{A})$ and $(\mathbf{Q}_{s}^{B}, \mathbf{Q}_{m}^{B})$. 
We then recombine them across prompts, e.g., $[\mathbf{Q}_{s}^{A}, \mathbf{Q}_{m}^{B}]$, and feed the mixed queries into the diffusion generator without additional training.
}

\hk{
As shown in Figure~\ref{fig:frankenstein}, the outputs often preserve the coarse layout or object configuration from prompt $A$ while adopting salient appearance attributes from prompt $B$. 
This recombination behavior, although not explicitly trained, suggests that IV-CoT learns a partially controllable structure-appearance separation in the latent query space. 
While the separation is not perfectly orthogonal, the results indicate that structural queries encode manipulable latent visual plans.
}

\section{Related Work}

% In stark contrast, our work bypasses these cumbersome RL scheduling and black-box morphing mechanisms. Instead, we introduce a highly transparent and geometrically grounded sketch information bottleneck. By regularizing the implicit reasoning process with an explicit topological lower bound, our method achieves native decoupled compositional generation without the training instability and latency overhead associated with RL-driven latent manipulation.

% \paragraph{Unified multimodal text-to-image generation.}
% Recent unified multimodal generators couple language understanding and visual synthesis within a shared model interface \citep{wu2025janus1,chen2025janus,wang2024emu3,xie2025show,xiao2025omnigen,zhou2025transfusion,ma2025janusflow,pan2025metaquery,wu2025openuni}. These models provide strong foundations for instruction-aware image generation, but their conditioning interface often remains a single entangled sequence in which layout, object identity, attributes, color, and texture are represented jointly. IV-CoT builds on query-conditioned MLLM-DiT generation, but reorganizes the conditioning interface into an ordered structural-to-semantic query cascade.

\paragraph{Explicit reasoning for image generation.}
With the great success of unified multi-modal large language models~\citep{zhou2025scale,lin2025toklip,xiao2026mindomni,xiao2026spatialedit}, broader language-model applications and multimodal learning~\citep{xiong2026mmformalizer,xiong2024dq,li2025ctr,li2024uncertaintyrag,hu2026emotion,chen2024tgca,chen2025mghft,sun2025divide}, and visual generation models~\citep{yang2026conceptguided,song20253sgen,song2026unialignment},
recent work improves compositional image generation by externalizing intermediate reasoning as text, layouts, scene plans, or visual drafts. LLM-based planning methods first produce scene descriptions, layouts, or other intermediate representations before invoking a diffusion generator \citep{feng2023layoutgpt,lian2023llm,galun2024generating,koch2025two}. Recent work has explored explicit reasoning for visual generation through pre-generation textual planning, RL-enhanced CoT, and post-generation reflection \citep{liao2025imagegen,guo2025can,tong2026delving,zhang2025reasongen,gu2025improving,zhang2025layercraft,li2025reflect,zhuo2025reflection,huang2025interleaving}. These methods externalize reasoning as textual plans, verification traces, visual drafts, or iterative refinement steps. IV-CoT instead keeps the intermediate visual plan inside latent queries and performs standard single-pass diffusion decoding at inference time.

\paragraph{Latent and continuous reasoning.}
Recent studies on latent or continuous reasoning suggest that intermediate deliberation can be carried by hidden states or soft thought tokens rather than natural-language rationales \citep{hao2024training,xu2025softcot,xu2025softcot++,ramji2026thinking}. Similar ideas have also been explored in multimodal understanding, where latent visual or multimodal tokens support reasoning without fully verbalizing intermediate steps \citep{pham2025multimodal,chen2025reasoning,wang2025mcot_survey,xiong2026mmformalizer}. Broader recent efforts explore latent representations for image generation via generation-time intervention \citep{mi2025milr,chen2026show,sun2026thinking}. IV-CoT instead structures the MLLM-DiT conditioning interface with structure-first queries, without intermediate decoding or test-time latent search.

% \paragraph{Structural guidance and compositional control.}
% Controllable text-to-image generation often injects explicit spatial conditions such as edges, sketches, depth maps, boxes, or grounding annotations into diffusion models \citep{zhang2023adding,mou2024t2i,li2023gligen}. Other methods improve compositional binding by analyzing or modifying cross-attention behavior \citep{tang2023daam,rassin2023linguistic,zhang2025rbind}. Unlike these approaches, IV-CoT uses sketches only as a training-time structural signal; at inference time, the structural guidance is internalized by learned structural queries, with no external sketch, layout, or intermediate visual state.

\section{Conclusion}

% We presented \textbf{IV-CoT}, a structure-first latent reasoning framework for text-to-image generation. By combining training-only sketch supervision with a structural-to-semantic query cascade, IV-CoT improves structure-aware prompt following on GenEval and T2I-CompBench while retaining single-pass inference without decoded intermediate sketches or textual rationales. Further analyses indicate that its structural queries encode recoverable visual plans and provide manipulable handles for latent structure-appearance control.

% We introduced \textbf{IV-CoT}, a structure-first latent reasoning framework for text-to-image generation that organizes conditioning queries into structural and semantic roles. With training-only sketch supervision and a structural-to-semantic query cascade, IV-CoT keeps visual planning inside latent representations without decoding intermediate sketches or textual rationales at inference time. Experiments on GenEval and T2I-CompBench show that IV-CoT improves structure-aware prompt following while maintaining high inference efficiency. Further analyses suggest that structural queries encode recoverable and manipulable visual plans, offering a practical direction for efficient and controllable structure-aware generation.

We introduce \textbf{IV-CoT}, a structure-first latent reasoning framework for text-to-image generation that organizes conditioning queries into structural and semantic roles. With training-only sketch supervision and a structural-to-semantic query cascade, IV-CoT keeps visual planning in latent representations without decoding intermediate sketches or textual rationales at inference time. Experiments and analyses on GenEval and T2I-CompBench show that IV-CoT improves structure-aware prompt following while maintaining high efficiency, with structural queries encoding recoverable and manipulable visual plans.

% \clearpage
% \newpage
\section*{Limitations}
This work focuses on structure-aware prompt following for text-to-image generation. First, IV-CoT is not specifically optimized for rendering readable text within images. Although sketch supervision encourages structural queries to capture contours, layouts, and object configurations, accurate scene-text rendering requires fine-grained character-level alignment, spelling consistency, and typography-aware supervision, which are not explicitly modeled in our current training objectives. Second, we mainly evaluate IV-CoT in text-to-image generation and have not explored image editing scenarios, such as localized editing, instruction-guided revision, or multi-turn refinement. Extending latent structural-to-semantic reasoning to text-aware generation and editing remains a promising direction for future work.

\bibliography{custom}

\clearpage
\newpage
\appendix

\section{Implementation Details}
\label{app:impl}

\paragraph{Backbone and query configuration.}
We instantiate IV-CoT on OpenUni-L, which combines a 2B InternVL3~\citep{zhu2025internvl3} MLLM with a 1.6B Sana diffusion generator~\citep{xie2025sana}. 
The model uses two groups of 256 visual queries, resulting in 512 conditioning queries after concatenation. 
The semantic queries are initialized from the pretrained OpenUni checkpoint, while the structural queries are initialized from the Stage-1 checkpoint. 
During Stage-2 training, the MLLM is frozen, and the Sana diffusion transformer, dual query inputs, and connector/projector modules are optimized. 
We set the structural regularization weight to $\lambda=0.3$.

\paragraph{Training data.}
We train on a combined dataset of 128,393 image-text pairs from BLIP3o, ShareGPT-4o-Image, and Echo-4o. 

\paragraph{Optimization.}
We train IV-CoT on NVIDIA A800 80GB GPUs using bfloat16 mixed precision. 
We use AdamW with learning rate $2\times10^{-5}$, $\beta=(0.9,0.95)$, weight decay 0.05, and gradient clipping at 1.0. 
The learning rate is linearly warmed up for the first 10\% of training steps and then decayed with a cosine schedule to $1\times10^{-7}$. 
Unless otherwise specified, we set the random seed to 42.

\section{Additional Generation Samples}
\label{appendix:additional_samples}

\begin{figure*}[htbp]
    \centering
    \includegraphics[width=\textwidth]{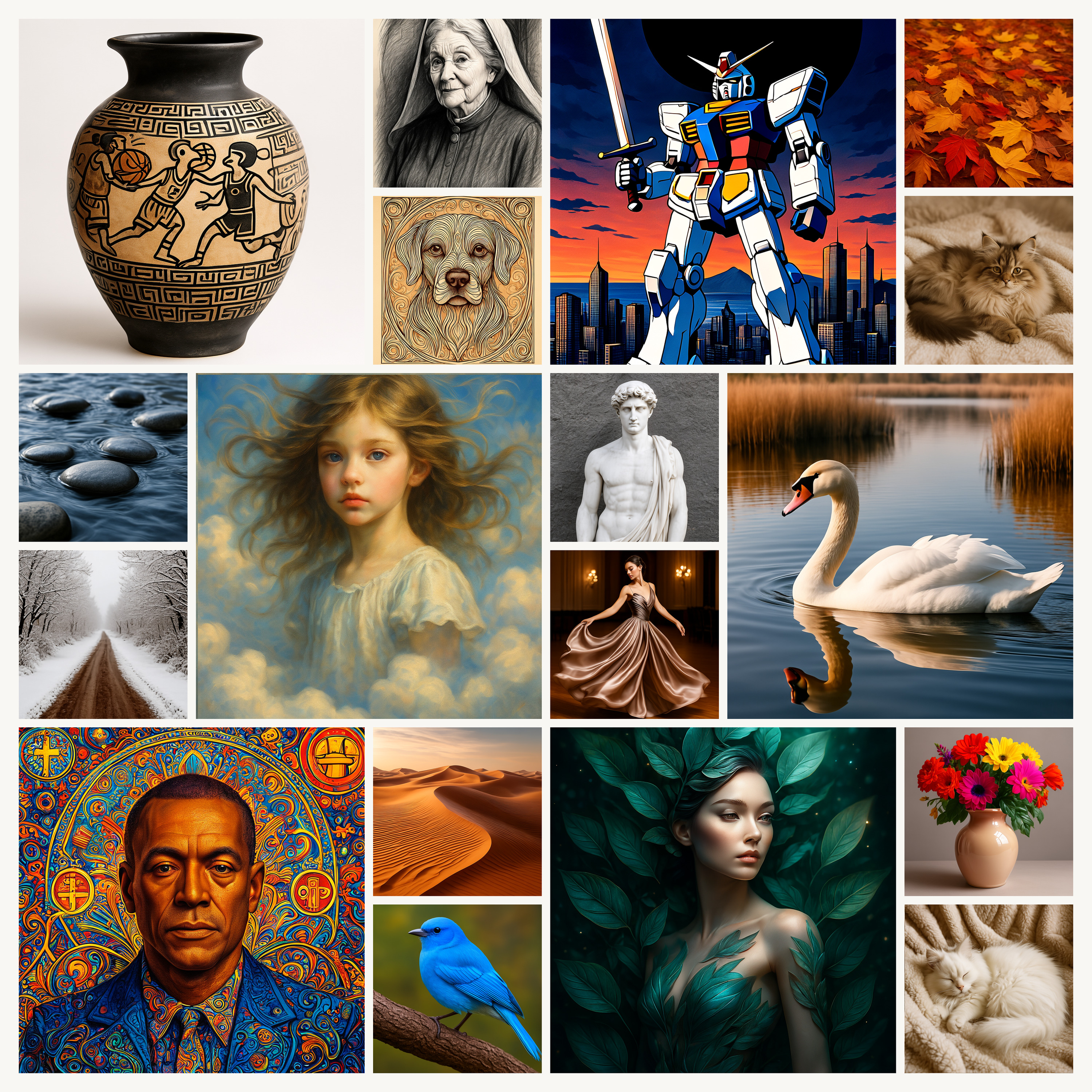}
    \caption{
    Additional qualitative samples generated by IV-CoT across diverse prompts.
    The examples cover objects, portraits, animals, natural scenes, and artistic styles,
    showing that the proposed structure-first latent reasoning framework maintains broad visual diversity while producing coherent image structures.
    }
    \label{fig:pic_show}
\end{figure*}

We provide additional generation samples from IV-CoT in Figure~\ref{fig:pic_show}. These examples cover diverse object categories, scenes, and visual styles, illustrating that IV-CoT maintains broad generation capability while preserving coherent visual structures.

\section{Latency Measurement}
\label{app:latency}

As efficiency is important for the generation process~\citep{lin2025quantization,lin2026duquant++,lin2026efficient,xing2025efficientllm,xia2025medrek,xu2026prune}, 
we measure latency on a single NVIDIA A800 80GB GPU with batch size 1. 
For each method, we report the average wall-clock inference time over 100 prompts, excluding model loading time. 
The time includes all steps required to produce the final image, including text processing, method-specific reasoning or intermediate generation, and final image synthesis.

\section{Attention Analysis}
\label{appendix:attention}

\paragraph{Relative attention proportion.}
To further examine how the diffusion generator allocates attention between structural and semantic queries during rendering, we compute the relative cross-attention proportion assigned to each query group. For each spatial latent position $p$, let $A(p,q)$ denote the cross-attention weight from position $p$ to query $q$. We define
\begin{align}
Z(p) &= \sum_{q\in \mathbf{Q}_{s}} A(p,q) + \sum_{q\in \mathbf{Q}_{m}} A(p,q), \\
r_s(p) &= \frac{\sum_{q\in \mathbf{Q}_{s}} A(p,q)}{Z(p)}, \\
r_m(p) &= 1-r_s(p),
\end{align}
where $r_s(p)$ and $r_m(p)$ denote the relative proportions assigned to structural queries $\mathbf{Q}_{s}$ and semantic queries $\mathbf{Q}_{m}$, respectively. Since the two query groups contain the same number of queries, this group-wise normalization is not confounded by query-group size. The maps should therefore be interpreted as relative attention allocations between the two query groups, rather than absolute attention magnitudes.

\begin{figure*}[t]
    \centering
    \includegraphics[width=\textwidth,height=0.86\textheight,keepaspectratio]{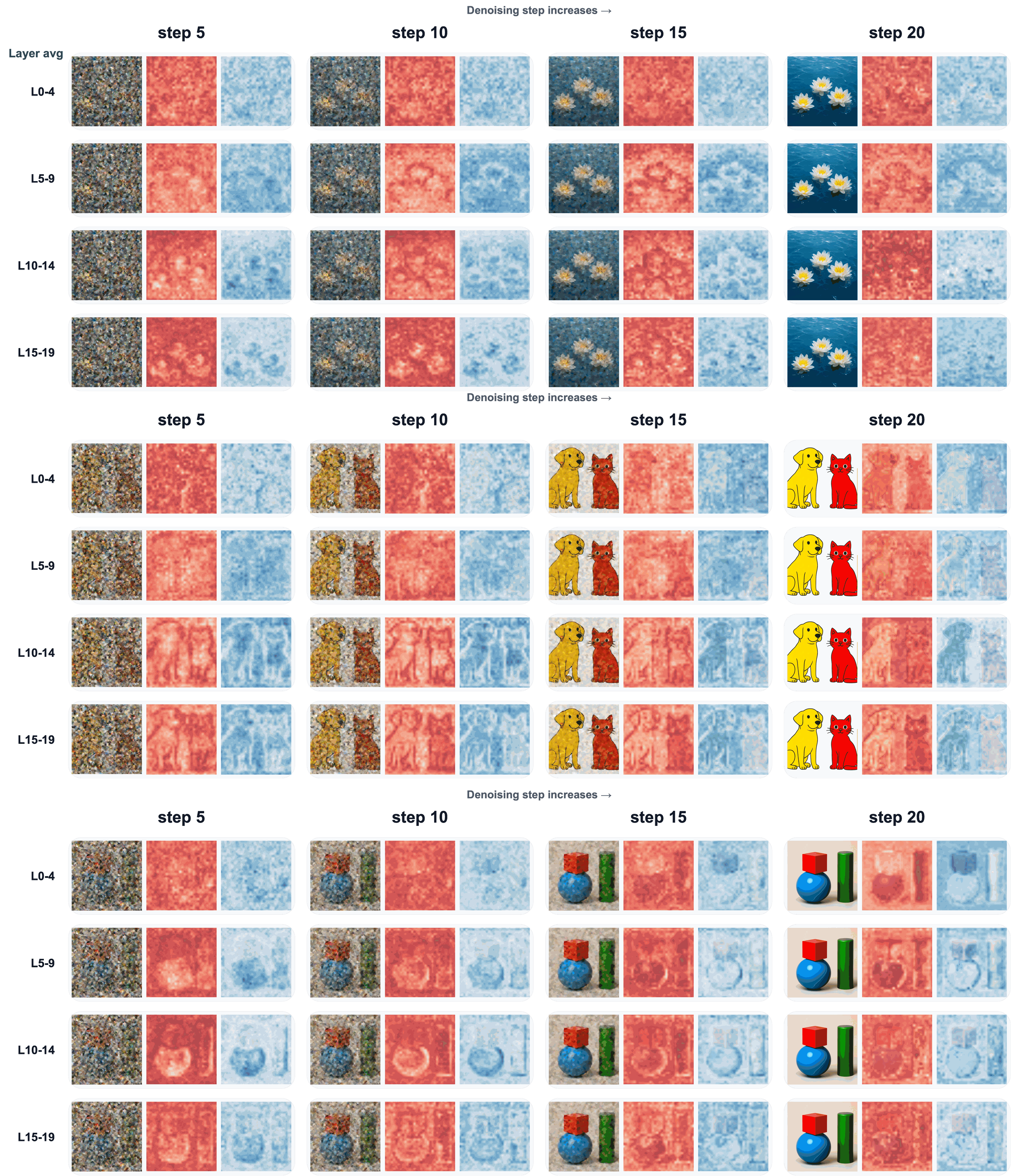}
    \caption{
Layer- and denoising-step-wise relative cross-attention proportion maps.
Columns show denoising steps 5, 10, 15, and 20, and rows show grouped diffusion-transformer layers.
Within each triplet, from left to right, we visualize the intermediate denoised image, the relative proportion assigned to structural queries $\mathbf{Q}_{s}$, and the complementary proportion assigned to semantic queries $\mathbf{Q}_{m}$.
At early denoising steps, structural-query patterns become more organized in deeper layers; at later steps, similar spatial patterns also emerge in shallower layers, suggesting progressive structure formation across denoising and depth.
}
    \label{fig:overall_heatmap}
\end{figure*}

\paragraph{Layer- and step-wise visualization.}
Figure~\ref{fig:overall_heatmap} expands the main attention analysis across denoising steps and diffusion-transformer layer groups. Columns correspond to increasing denoising steps, and rows group consecutive layers. Within each triplet, from left to right, we show the intermediate denoised image, the relative attention proportion assigned to structural queries $\mathbf{Q}_{s}$, and the complementary proportion assigned to semantic queries $\mathbf{Q}_{m}$.

We observe a layer--step interaction. At early denoising steps, when the latent image state is still noisy, structural patterns are more visible in deeper layers, suggesting that deeper layers aggregate global information to recover coarse layouts. As denoising progresses, similar spatial patterns also appear in shallower layers, indicating that structural queries increasingly align with local object regions once coarse structures have emerged. The semantic-query maps show complementary and often more diffuse allocation patterns, suggesting soft functional specialization rather than a hard separation.

\section{Use of AI Assistants}

The authors used AI assistants for language polishing, wording suggestions, and submission-form preparation. All technical content, experiments, analyses, claims, and final text were reviewed and verified by the authors.

% This is an appendix.

\end{document}